\documentclass{da2pl2026}
\usepackage{times}
\usepackage{graphicx}
\usepackage{latexsym}
\usepackage{eurosym}
\usepackage{amsmath,amsthm,amssymb,stmaryrd,pifont,wasysym}
\usepackage{dsfont}
\usepackage{pstricks,pst-plot,pst-text,pst-tree,pst-eps,pst-fill,pst-node,pst-math}
\usepackage{microtype}

\usepackage{booktabs, multirow}
\usepackage{algorithm,algorithmic}

\ecaisubmission 

\begin{document}

\title{Bayesian Preference Elicitation: Human-In-The-Loop Optimization of An Active Prosthesis}

\author{
Sophia Taddei$^{1,2}$, 
Wouter Koppen $^{3}$, 
Eligia Alfio$^{2}$, 
Stefano Nuzzo$^{1,4}$, 
Louis Flynn$^{1,4}$, 
María Alejandra \\ Díaz$^{2}$, 
Sebastian Rojas Gonzalez$^{3,5}$, 
Tom Dhaene$^{3}$,
Kevin De Pauw$^{2}$,
Ivo Couckuyt$^{3}$, 
Tom Verstraten$^{1,4}$ 
}


\maketitle
\bibliographystyle{ecai2012}

\begin{abstract}Tuning active prostheses for people with amputation is time-consuming and relies on metrics that may not fully reflect user needs. We introduce a human-in-the-loop optimization (HILO) approach that leverages direct user preferences to personalize a standard four-parameter prosthesis controller efficiently. Our method employs preference-based Multiobjective Bayesian Optimization that uses a state-or-the-art acquisition function especially designed for preference learning, and includes two algorithmic variants: a discrete version (\textit{EUBO-LineCoSpar}), and a continuous version (\textit{BPE4Prost}). Simulation results on benchmark functions and real-application trials demonstrate efficient convergence, robust preference elicitation, and measurable biomechanical improvements, illustrating the potential of preference-driven tuning for user-centered prosthesis control.
\end{abstract}


\setcounter{footnote}{1}
\footnotetext[1]{Federated Labs Robotics $\&$ AI, Vrije Universiteit Brussel.}
\footnotetext[2]{Human Physiology and Sports Physiotherapy Research Group, Vrije Universiteit Brussel.}
\footnotetext[3]{IDLab, Ghent University -- imec.}
\footnotetext[4]{Robotics $\&$ Multibody Mechanics Research Group, Flanders Make.}
\footnotetext[5]{Corresponding author: sebastian.rojasgonzalez@ugent.be}

\section{Introduction}
\label{sec:intro}

Robotic prostheses are designed to replace a missing limb and restore mobility after an amputation. However, passive prostheses cannot provide the necessary mechanical energy that biological muscles normally generate~\cite{babivc2021challenges}. Therefore, the field is advancing into active prostheses that use electric motors to generate mechanical energy, enabling powered assistance. However, they rely on complex control systems that define the assistance of the device through adjustable parameters. Normally, these parameters must be individually tuned for each user by the prosthetist~\cite{felt2015body}. This process often relies on heuristic trial-and-error methods, which are time-consuming and make it harder to adopt powered prostheses in clinical practice.~\cite{simon2014configuring,gehlhar2023review}. To overcome this challenge, frameworks like Human-in-the-Loop Optimization (HILO) \cite{slade2024human}, and Bayesian Preference Elicitation (BPE) \cite{huber2025bayesian}, have emerged as promising solutions for efficient, user-centered parameter tuning. 

HILO is a methodology that automatically adjusts control parameters to improve user performance~\cite{diaz2022human}. Objective functions in HILO are biomechanical or physiological measures of user performance, such as energy expenditure or gait symmetry (see e.g., exoskeleton studies~\cite{tucker2020preference,slade2024human, zhang2017human, song2021optimizing}, and prosthesis applications ~\cite{clites2021understanding, welker2021shortcomings, alili2023novel, wen2018robotic}). However, the challenges in prosthetic tuning differ substantially from those in exoskeletons due to the distinct demands of prosthesis use~\cite{welker2021shortcomings}. In particular, prostheses place greater emphasis on stability, comfort, and coordination, which in turn make the design of objective functions and tuning process challenging~\cite{welker2021shortcomings}. In this context, user preference provides a natural objective, as it implicitly integrates \emph{multiple} physiological and biomechanical factors and objectives without requiring strong assumptions about gait performance and user need~\cite{tucker2020preference}. 

Optimizing subjective, user-centered objectives aligns well with BPE, where feedback is collected through relative comparisons rather than absolute scores. This is because numerical ratings are often unreliable because people change their preferences over time, shift their internal rating scale, and are easily influenced by contextual biases~\cite{ingraham2023leveraging}. In contrast, humans tend to be more consistent when expressing ordinal preferences (e.g., ``$A$ feels better than $B$'')~\cite{brochu2010tutorial}.
BPE formalizes this idea by introducing a latent utility function $u:\mathcal{X}\to\mathbb{R}$ that represents the user’s internal utility model. This utility is typically modeled with a Gaussian Process (GP) and combined with probabilistic choice models such as the logistic likelihood~\cite{chu2005preference, nguyen.etal:2021}, enabling learning directly from pairwise comparisons without requiring explicit measurement of any gait metrics. Related approaches appear under different names in literature: Preferential Bayesian Optimization (PBO), Bayesian Optimization with Preference Elicitation/Exploration (BOPE), or interactive BO. In this work, we adopt the umbrella term \emph{BPE} to refer to these preference-driven Bayesian optimization methods.

A key challenge in BPE is selecting informative comparisons. Unlike classical BO, where the acquisition function operate on an observable scalar objective $f(x)$~\cite{couckuyt2022bayesian,garnett2023bayesian}, BPE acquisition functions must reason over relative information. Early work introduced acquisition objectives directly based on the GP posterior~\cite{gonzalez2017preferential}. More recently, the \emph{Expected Utility of the Best Option} (EUBO) and its batch extension $q$EUBO \cite{astudillo2023qeubo} provide a principled, decision-theoretic criterion tailored to evaluate a candidate set based on the expected utility of its best element. These methods have shown strong empirical performance in both simulation and human-in-the-loop experiments, both for single- and multi-objective problems \cite{lin2022preference,huber2025bayesian}. Because each comparison requires both physical effort and careful judgment from the user, query efficiency is particularly important in prosthesis tuning.

In wearable robotics, user preference has been elicited using BPE approaches~\cite{tucker2020preference, tucker2020human,arens2025preference} as well as evolutionary strategies such as covariance matrix adaptation evolution strategy~\cite{lee2023user}. For instance, Arens et al.\cite{arens2025preference} used a PBO framework incorporating Just-Noticeable Difference information and an Upper-Confidence Bound acquisition function to tune two parameters of a soft exosuit~\cite{arens2025preference}, while Tucker \textit{et al.}\cite{tucker2020preference} introduced \textit{LineCoSpar}, a BPE method based on Thompson Sampling (TS), to tune six gait parameters of an assistive exoskeleton.

In this work, we propose a progressive approach to BPE for prosthesis tuning. Inspired by \cite{tucker2020preference}, we first introduce a discretized variant (denoted \textit{EUBO-LineCoSpar}), simplifying the parameter space to reduce computational complexity while validating feasibility. The approach replaces TS (as done in \cite{tucker2020preference}) with EUBO, improving sample efficiency and convergence, particularly in higher-dimensional settings. Building on recent results in BPE for multiobjective optimization \cite{huber2025bayesian}, we then propose a framework, denoted \textit{BPE4Prost}, to a continuous parameter space using a Logit likelihood enabling richer exploration and improved robustness under noisy feedback. By embedding a principled acquisition strategy designed especially for Human-in-the-Loop optimization for prosthesis control personalization, we observe faster convergence and more reliable parameter tuning than existing procedures. Both proposed algorithms are evaluated in simulation and further assessed in a real-life prosthesis control application.

\section{Bayesian Preference Elicitation (BPE)}
\label{sec:section11}

In this section, we introduce the fundamentals of the BPE framework; for more details and references we refer to \cite{huber2025bayesian,astudillo2023qeubo,nguyen.etal:2021,lin2022preference}.

\subsection{Problem Definition}
Let $x \in \mathcal{X} \subset \mathbb{R}^d$ denote a configuration or parameter vector (e.g., tunable prosthesis control parameters). The goal is to identify a configuration that maximizes the user's latent utility 
$u : \mathcal{X} \to \mathbb{R}$:
\begin{equation}
    x^\star \in \arg\max_{x \in \mathcal{X}} u(x).
\end{equation}

Since the utility function $u(x)$ cannot be observed directly, the only feedback available is the user's pairwise preference between two proposed configurations. At iteration $n$, the algorithm presents a pair $(x_{n,1}, x_{n,2})$, and the user reports which one is preferred:
\begin{equation}
\label{eq:r}
r_n = 
\begin{cases}
1, & \text{if } x_{n,1} \succ x_{n,2}, \\
2, & \text{if } x_{n,2} \succ x_{n,1}.
\end{cases}
\end{equation}
These comparisons reveal the ordering of $u(x_{n,1})$ and $u(x_{n,2})$, possibly corrupted by the judgment noise of the user.

BPE begins with an initialization phase consisting of $N_{\text{init}}$ random pairwise comparisons to build an initial surrogate model. Then it proceeds by repeatedly: (i) selecting a comparison pair; (ii) obtaining a noisy preference from the user; and (iii) updating a probabilistic model of the latent utility. After $N$ queries, the collected dataset is
\[
\mathcal{D} = \{(x_{n,1}, x_{n,2}, r_n)\}_{n=1}^N,
\]
and the set of all evaluated designs is
\[
\mathcal{W} = \{x_{n,i} \mid n = 1,\dots,N,\ i \in \{1,2\} \}.
\]

\subsection{Probabilistic Modeling}
We place a GP prior on the latent utility:
\begin{equation}
    u(x) \sim \mathcal{GP}(\mu(x), K(x,x')),
\end{equation}
where $\mu(\cdot)$ and $K(\cdot,\cdot)$ denote the mean function and kernel \cite{rasmussen2005gpml, frazier2018tutorial}. 

Because utilities are not observed directly, we use a Bradley-Terry logistic likelihood~\cite{chu2005preference} to connect pairwise comparisons to utility differences. The logistic likelihood is defined as follows:
\begin{equation}
\label{eq:bt}
\begin{split}
    P(r_n = 1 \mid u) &= 
    \sigma_\lambda\!\left(u(x_{n,1}) - u(x_{n,2})\right) \\
    &= \frac{1}{1 + \exp\!\left(-\frac{u(x_{n,1}) - u(x_{n,2})}{\lambda}\right)},
\end{split}
\end{equation}
where $\lambda > 0$ controls the noise level in the user's judgments. Assuming independent observations, the joint likelihood is
\[
\mathcal{L}(\mathcal{D} \mid u) 
= \prod_{n=1}^N P(r_n \mid u).
\]
where $r_n = {1,2}$, just as denoted in Equation \ref{eq:r}. 

Because the logistic likelihood is non-conjugate to the GP prior, the posterior $P(u \mid \mathcal{D})$ is analytically intractable. 
Following standard BPE implementations in BoTorch~\cite{balandat2020botorch}, we use the Laplace approximation~\cite{cheikh2010method} to obtain a Gaussian posterior approximation $\mathcal{N}(\mu', \Sigma')$. 
Other approximation strategies, such as variational inference~\cite{nguyen.etal:2021}, have also been explored in the literature, but are not considered here.
The resulting Gaussian posterior enables efficient Monte Carlo sampling, which is required for efficiently evaluating the acquisition functions used in BPE.

\subsection{Preference elicitation strategy}
Pairwise preferences provide only relative information, which makes classical BO acquisition functions (e.g., Expected Improvement) unsuitable. Instead, BPE uses acquisition functions specifically designed for preference-driven settings. We adopt EUBO \cite{lin2022preference, astudillo2023qeubo,huber2025bayesian}, which is particularly suitable in this setting for three reasons: (i) it targets high-utility configurations rather than utility differences, aligning with the underlying goal; (ii) the learned utility encodes \emph{multiobjective} preferences implicitly, avoiding the need for the human to understand trade-offs in objective space; and (iii) prior work has demonstrated strong empirical performance for up to nine objectives \cite{huber2025bayesian}. For a batch of $q$ candidates, the $q$EUBO acquisition function is
\[
q\mathrm{EUBO}(\{x_1,\dots,x_q\})
=
\mathbb{E}_{n-1}\!\left[\max_{i=1,\dots,q} u(x_i)\right],
\]
where the expectation is taken with respect to the posterior after $n{-}1$ preference queries. The next batch is chosen by
\begin{equation}
X_n \in \arg\max_{X \in \mathcal{X}^q} q\mathrm{EUBO}_{n-1}(X).
\label{eq:qeubo_selection}
\end{equation}

$q$EUBO favors batches that are likely to contain at least one high-utility design while still exploring uncertain regions. 

\section{Algorithms}
\label{sec:section12}
This section presents the two variants of the algorithm, both implemented within the BPE framework. To ensure reproducibility and support to other scientific communities, we make our implementation details and simulation code publicly available online\footnote{https://github.com/sophtddi/Preference-HILO-Prosthesis}$^,$\footnote{https://github.com/WKoppen/Simulations-VUB/tree/main}.

\subsection{\textit{EUBO-LineCoSpar}}
\textit{EUBO-LineCoSpar} constitutes a discrete and interpretable formulation, designed to facilitate rapid experimentation, to give a first understanding of user preference in the context of wearable robotics control. The discrete formulation uses $N_{samples}$ points per dimension to simplify the problem from a control perspective.

\begin{algorithm}[h] 
\caption{\textit{EUBO-LineCoSpar}}
\label{alg:pref_bo} 
\scriptsize 
\begin{algorithmic}[1] 
    \STATE Initialize dataset $\mathcal{D} \leftarrow \mathcal{D}_0$, visited set $\mathcal{W} \leftarrow \mathcal{W}_0$ 
    \STATE Sample initial pair $(x_{0, 1}, x_{0, 2})$ and query preference $x_{0,\text{pref}}$ 
    \FOR{$n = 1, \dots, N$}
        \STATE Update $\mathcal{D}$, $\mathcal{W}$, $x_{\text{pref}}$, and compute GP posterior
        \STATE Generate a random line $L_k$ passing through $x_{\text{n-1,pref}}$ 
        \STATE Define candidate set $V_n = L_n \cup \mathcal{W}$ 
        \STATE Select challenger $x_{n, \text{chal}} \in V_n$ using acquisition function 
        \STATE Query user preference between {$x_{n-1,\text{pref}}$, $x_{n, \text{chal}}$}
        \STATE Update $x_{n, \text{pref}}$,
            \IF{$x_{n, \text{pref}} = x_{n-1, \text{pref}}$ for $N_{\text{stop}}$ iterations} 
                \STATE \textbf{break} 
            \ENDIF 
    \ENDFOR 
    \STATE \textbf{return} $\hat{x} = x_{N, \text{pref}}$ 
\end{algorithmic} 
\end{algorithm}

The method follows the general structure of the \textit{LineCoSpar} algorithm proposed by Tucker \textit{et al.}~\cite{tucker2020human}, adapted to our application domain. In contrast to \textit{LineCoSpar}, which relies on TS to select candidate challengers, \textit{EUBO-LineCoSpar} employs EUBO to select the challenger configuration. This choice enables a more efficient optimization of the latent utility function inferred from preference feedback.

The implemented algorithm is detailed in Algorithm~\ref{alg:pref_bo}. At each iteration $n$, the algorithm presents to the user two configurations in a random order: (1) $x_{n-1,\text{pref}}$, the configuration selected as the preferred one at previous iteration; (2) a newly proposed configuration, $x_{n,\text{chal}}$, selected as the point maximizing EUBO on a candidate set formed by the union of a random line passing through $x_{n-1,\text{pref}}$ and the set of previously visited configurations $\mathcal{W}$. The algorithm is designed to terminate early if the user maintains the same $x_{\text{pref}}$ for $N_{\text{stop}}$ consecutive iterations. The algorithm’s estimate of the user’s preferred configuration, $\hat{x}$, taken as the final preferred configuration, $x_{N, \text{pref}}$.

\subsection{Bayesian Preference Elicitation for Prosthesis Optimization: BPE4Prost}
\textit{BPE4Prost} extends the \textit{EUBO-LineCoSpar} framework to a continuous design space by directly optimizing the $q$EUBO acquisition function over the entire domain. At each iteration, the algorithm selects three candidate configurations that jointly maximize the $q$EUBO criterion (Eq.~\ref{eq:qeubo_selection}). A pairwise preference query is first issued between the two leading candidates. If the user expresses no preference, a new query is formed between the first and the third candidate. If indifference persists, the first candidate is sequentially compared with randomly generated configurations until a preference is obtained. This procedure ensures informative preference feedback while maintaining exploration over the full design space. 

\begin{algorithm}[h] 
\caption{\textit{BPE4Prost}} \label{alg:bope2_eubo} 
\scriptsize 
\begin{algorithmic}[1] 
    \STATE Initialize dataset $\mathcal{D} \leftarrow \mathcal{D}_0$, visited set $\mathcal{W} \leftarrow \mathcal{W}_0$ 
    \STATE Define initial GP posterior 
    \FOR{$n = 1, \dots, N$} 
        \STATE Compute EUBO acquisition function 
        \STATE Select candidate triplet $\{x_{n,1}, x_{n,2}, x_{n,3}\}$ maximizing EUBO 
        \STATE Query user preference between $\{x_{n,1}, x_{n,2}\}$             \IF{preference observed} 
                \STATE Update $\mathcal{D}$, $\mathcal{W}$ 
            \ELSE 
                \STATE Query user preference between $\{x_{n,1}, x_{n,3}\}$ 
                \WHILE{no preference observed} 
                    \STATE Sample random configuration $x_{\text{random}}$ 
                    \STATE Query user preference between $\{x_{n,1}, x_{\text{random}}\}$ 
                \ENDWHILE 
                \STATE Update $\mathcal{D}$, $\mathcal{W}$ 
            \ENDIF 
        \STATE Compute GP posterior 
        \STATE Compute $\hat{x}_n = \arg\max_x \mu_N(x)$ 
        \IF{no improvement in $\hat{x}_n$ for $N_{\text{stop}}$ iterations} 
            \STATE \textbf{break} 
        \ENDIF 
    \ENDFOR 
    \STATE Compute mean GP posterior $\mu_{N}(x)$ 
    \STATE \textbf{return} $\hat{x} = \arg\max_x \mu_{N}(x)$ 
\end{algorithmic} 
\end{algorithm}\vspace{1em}

The estimate of user preferred configuration, $\hat{x}_n$, is determined at each iteration $n$ as the maximizer of the posterior mean of the GP, $\mu_n(x)$. To ensure practical convergence, the algorithm incorporates a stopping criterion based on stability of the preferred configuration: if the maximizer remains within a $\pm 10\%$ neighborhood of the same point in the parameter space for $N_{\text{stop}}$ consecutive iterations, the optimization process is terminated. This threshold is chosen to reflect that differences smaller than 10\% are typically imperceptible to the user, and thus further queries would not provide meaningful information for preference learning. The algorithm’s estimate of the user’s preferred configuration, $\hat{x}$, is then taken as the maximizer of the final posterior mean, i.e. $\arg\max_x \mu_N(x)$. 

Note that \textit{EUBO-LineCoSpar} uses a Probit likelihood and an exponential kernel. In contrast, \textit{BPE4Prost} adopts the logistic likelihood and a Matérn kernel ($\nu = 5/2)$, following Huber \textit{et al.} \cite{huber2025bayesian}. This choice provides more flexible noise modeling through the parameter $\lambda$ and imposes a smoother prior over the latent utility function, which is typically more appropriate in preferential optimization settings.

\section{Simulation evaluation}
\label{sec:section2}

This section evaluates the empirical performance of the proposed algorithms. Section~\ref{sec:simulation_benchmark} describes the simulation setup, and Section~\ref{sec:simulation_results} presents the results and compares the proposed methods to a random sampling baseline. 

\begin{figure*}[ht!]
    \centering
    \includegraphics[width=0.9\textwidth]{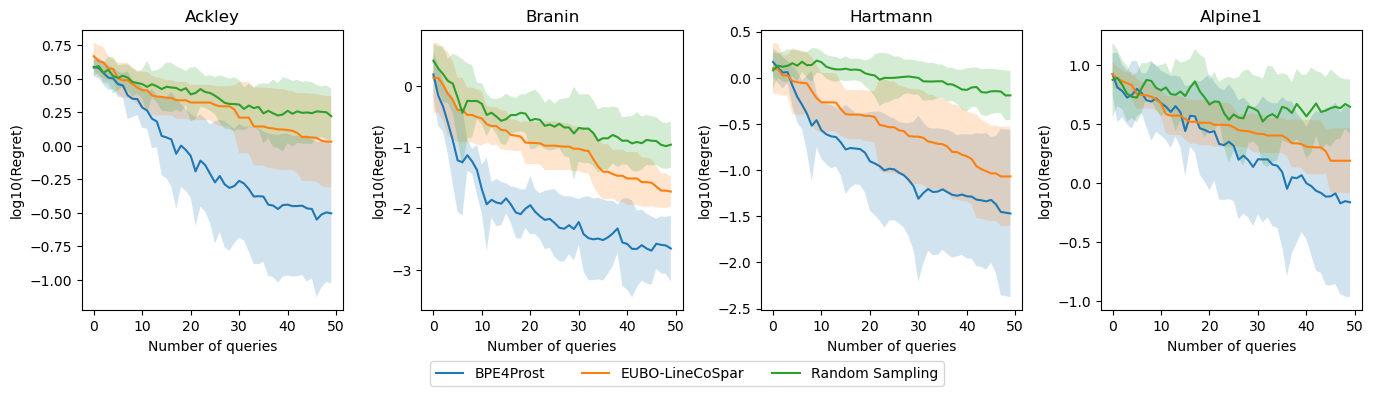}
    \vspace{-15pt}
    \caption{ 
Mean $\log_{10}$ simple regret over 11 runs for all benchmark functions. 
Each algorithm was run for 50 iterations. Shaded regions indicate the standard error. 
Across both the 2D problem (Branin) and the 4D problems (Ackley, Alpine1, Hartmann), 
BPE4Prost consistently achieves lower regret compared to the other methods.}
    \label{fig:simulations}
\end{figure*}

\subsection{Simulation Setup}
\label{sec:simulation_benchmark}
We first evaluate all methods on standard continuous optimization benchmarks implemented in BoTorch \cite{balandat2020botorch}: Branin (2D), Ackley (4D), Alpine1 (4D), and Hartmann (4D). These functions span different levels of multimodality and dimensionality, and they represent the utility of an arbitrary human that the algorithm must learn. Each benchmark serves as a ground-truth latent utility $u:\mathcal{X}\to\mathbb{R}$, where a configuration $x_1$ is preferred to $x_2$ whenever $u(x_1) > u(x_2)$. We compare the two proposed methods, \textit{EUBO-LineCoSpar} and \textit{BPE4Prost}, against a random sampling baseline. In the random baseline, pairs of candidates are formed from quasi-random Sobol sequence samples drawn i.i.d.\ over $\mathcal{X}$.
\vspace{3pt}
\\
\textbf{Initialization and budget.}
To reflect the limited evaluation budget typical of prosthesis tuning and similar applications, each run typically begins by sampling $2(2d{+}1)$ points from a quasi-random Sobol sequence \cite{balandat2020botorch}, which are paired to yield the $N_{init}$ initial comparisons. Optimization then proceeds for an additional $N = 50$ preference queries, for a total of $2d{+}1 + 50$ comparisons per run, which is unrealistically high (i.e., we expect to converge much earlier in number of interactions). 
\vspace{3pt}
\\
\textbf{Surrogate modeling and common settings.}
Both \textit{BPE4Prost} and the random sampling baseline uses a logistic likelihood and Matérn kernel. For \textit{EUBO-LineCoSpar}, we adopt the Probit likelihood and Exponential kernel used in the original \textit{LineCoSpar} implementation~\cite{tucker2020preference}.
Hyperparameters are re-estimated after each query using the accumulated dataset $D_n$. To ensure a fair comparison, the set of initial configurations is the same for every method.
\vspace{3pt}
\\
\textbf{Domain discretization.}
Because \textit{LineCoSpar }operates on a discretized parameter domain, we implement \textit{EUBO-LineCoSpar} on a regular grid with $N_{samples}= 51$ points per dimension. This ensures a dense yet tractable search space.
\vspace{3pt}
\\
\textbf{Acquisition optimization.}
At each iteration $n$, every method select the next pair (or batch) by maximizing their acquisition function over the search space $\mathcal{X}$. For simulations, a batch size of $q=2$ was adopted, as always a preferred configuration could be chosen. Optimization uses multi-start L-BFGS-B following standard BoTorch practice~\cite{balandat2020botorch}.
\vspace{3pt}
\\
\textbf{Noise model.}
Simulations assume noiseless preferences so that observed comparisons always agree with the ground-truth ordering. Although logistic-likelihood preference models are robust to moderate noise in practice \cite{huber2025bayesian}, the noiseless setting allows a clearer comparison of optimization performance.
\vspace{3pt}
\\
\textbf{Evaluation metric.}
We report (log-scaled) simple regret with respect to the ground-truth utility $u$. Let $x^* \in \arg\max_{x\in\mathcal{X}} u(x)$ denote the global maximizer. After $n$ comparisons, the algorithm recommends $\hat{x}_n \in \arg\max_{x\in\mathcal{X}} \mathbb{E}[u(x)\mid \mathcal{D}_n]$, i.e., the maximizer of the surrogate’s posterior mean.
We define the simple regret as
\begin{equation}
    \mathrm{regret}_n = u(x^*) - u(\hat{x}_n),  
    \label{eq:regret}
\end{equation}
where lower values indicate faster convergence. We apply a $\log_{10}$ transformation when reporting results for improved visualization.

\subsection{Simulation Results}
\label{sec:simulation_results}

Figure~\ref{fig:simulations} reports the mean $\log_{10}(\mathrm{regret}_n)$ as a function of the number of pairwise queries (as defined in Equation \ref{eq:regret}), with shaded regions indicating the standard error over $11$ trials. Across all benchmark functions, the results indicate that BPE can efficiently identify high-utility regions in latent objective landscapes. \textit{BPE4Prost} consistently achieves the lowest regret, outperforming both random sampling and \textit{EUBO-LineCoSpar}. Its rapid initial decrease in regret indicates that it leverages the GP posterior effectively to focus on promising areas of the search space. The only exception is the Alpine1 function, whose shallow landscape and numerous local optima make rapid convergence more challenging. Even in this case, \textit{BPE4Prost} maintains a clear advantage over the other methods, especially after $50$ iterations.

It is interesting to observe the gap between the Bayesian strategies and random sampling widens with increasing dimensionality. On the 4D Ackley, Hartmann and Alpine1 functions, random sampling stagnates due to the large search volume, while the GP-based methods systematically eliminate suboptimal regions as posterior uncertainty decreases. \textit{EUBO-LineCoSpar} improves over random sampling but converges more slowly, reflecting its structural constraint of exploring only along one-dimensional lines through the current best configuration.

\begin{figure}[h]
    \centering
    \includegraphics[width=0.45\textwidth]{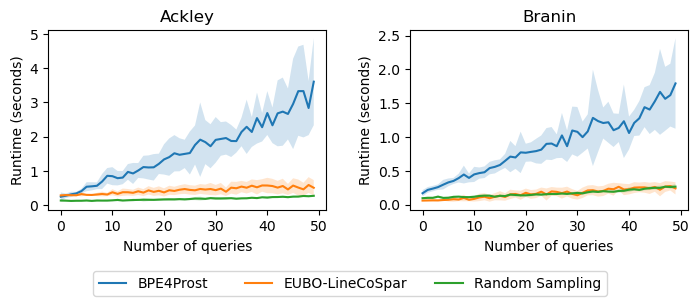}
    \vspace{-15pt}
    \caption{ 
Runtime comparison of the three algorithms on Ackley (4D) and Branin (2D). BPE4Prost scales approximately linearly with the number of queries, while EUBO-LineCoSpar remains near-constant due to its line-restricted evaluations. }
    \label{fig:runtimes}
\end{figure}

\begin{figure*}[!ht]
    \centering
    \includegraphics[width=0.9\textwidth]{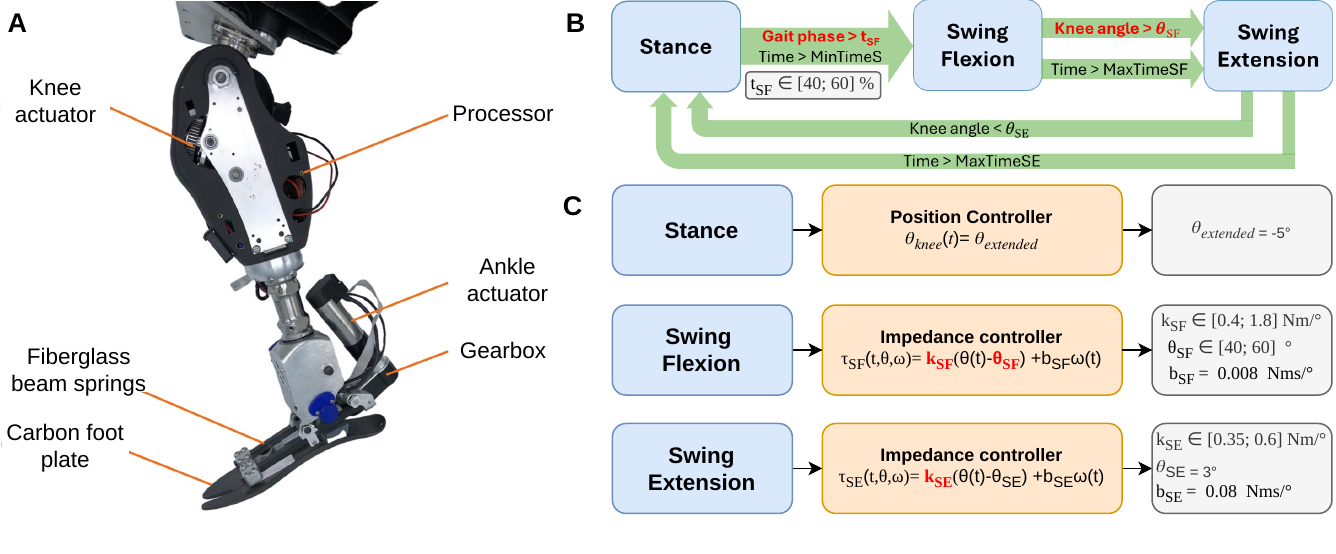}
    \vspace{-20pt}
    \caption{ 
    [A] Active prosthetic knee system used in this study.
    [B] High-level control framework based on a finite state machine (FSM) with three states: Stance (S), Swing Flexion (SF), and Swing Extension (SE). Threshold rules between states are indicated with tunable parameters written in red.
    [C] Mid-level control framework. A position controller is used in S, while impedance controllers are employed in SF ans SE. Tunable parameters are indicated in red.}
    \label{fig:prosthesis_and_control}
\end{figure*}

Figure~\ref{fig:runtimes} compares the computational costs of the methods. The runtime of \textit{BPE4Prost} scales approximately linearly with the number of queries, consistent with standard BO methods. In contrast, \textit{EUBO-LineCoSpar} exhibits nearly constant runtimes and is only slightly slower than random sampling, owing to its line-restricted, discretized evaluation strategy. Although this efficiency is appealing in resource-constrained settings, excessive comparisons may increase cognitive load on users in real experiments. In this context, the query efficiency of \textit{BPE4Prost} is preferable despite its \emph{negligible} higher computational cost.

\section{Human subject experiment}
\label{sec:results}
In this section, the algorithms are tested to find the optimal set of control parameters of the CYBERLEGs X-Leg (BruBotics, VUB, Belgium), a dual-actuated transfemoral prosthesis (Fig.~\ref{fig:prosthesis_and_control}A), during walking at a fixed, self-selected speed on a treadmill.

\subsection{Tunable Control Framework}
\label{sec:human_exp}

The control of the active device is hierarchically organized~\cite{diaz2022human}. A high-level controller manages the transition between the different gait state (e.g., stance or swing). A mid-level controller computes the required control command within each state. And, a low-level controller translates the control command in the required current output.
\vspace{3pt}
\\
\textbf{High-level control.} The Finite State Machine (FSM) defines threshold-rules to allow the transition from one gait state to another, based on kinematics and kinetics data~\cite{diaz2022human}. In our controller, we defined 3 gait states: Stance (S), Swing Flexion (SF), Swing Extension (SE), for which the transition rules are depicted in Fig.~\ref{fig:prosthesis_and_control}B.

The transition from S to SF is governed by a minimum duration ($MinTimeS$) to avoid early mistransition, and on a gait phase variable ($\Phi$) that captures the temporal evolution of walking independent of speed~\cite{diaz2022human}. $\Phi$ is computed from the acceleration and gyro measured by onboard IMUs on the user’s residual limb as presented by Quintero \textit{et al.}\cite{quintero2018continuous}. The threshold value on $\Phi$, $t_{SF}$, is tunable as beginning of SF depends on gait speed and step length~\cite{grieve1966relationships}, making it a meaningful user-dependent parameter to optimize. 

Transition from SF to SE is controlled by a maximum duration ($MaxTimeSF$) as a safety threshold, and by a condition on the knee angle $\theta$: when $\theta$ reaches the targeted knee flexion angle $\theta_{SF}$, the knee goes to extension. Maximum knee flexion angle depends also on gait speed and is a user-dependent variable~\cite{mentiplay2018lower}, making it another meaningful parameter to tune to each user preference. 

Transition from SE to S indicates the beginning of a new gait cycle. The new step starts when the knee is fully extended or after a maximum duration (MaxTimeSE) to ensure user safety. 
\vspace{3pt}
\\
\textbf{Mid-level control.} Within each state, impedance control strategies are typically used for prosthetic control, as natural walking have been shown to behave like variable impedance controllers and offer safer interaction with the environment~\cite{gehlhar2023review}. For safety considerations due to the hardware, a position controller that enforces a fully extended knee is implemented during S, whereas impedance controllers are used during SF and SE as shown in Fig.~\ref{fig:prosthesis_and_control}C. Impedance control law is given by: 
\begin{equation}
\tau_i(\theta(t), \dot{\theta}(t)) = k_i \left(\theta(t) - \theta_i\right) + b_i \dot{\theta}(t).
\label{eq:impedance}
\end{equation}

The impedance controller are defined for the phases $i \in \{\mathrm{SF}, \mathrm{SE}\}$, $\theta$ denotes the knee angle and $\dot{\theta}$ its angular velocity, both measured using embedded encoders. The impedance parameters consist of the stiffness $k_i$, the equilibrium angle $\theta_i$, and the damping coefficient $b_i$.

In our controller, we set $b_{SF}$, $b_{SE}$, and $\theta_{SE}$, to fixed values reported in the literature~\cite{sup2008design, wang2013design} and adapted to the device characteristics, to reduce the dimensionality of the tuning problem. Consequently, only $k_{SF}$, $\theta_{SF}$, and $k_{SE}$ are treated as user-dependent tunable parameters.
\\
\textbf{Tunable parameters.} The control system is parameterized by four primary tunable variables: (1) $t_{SF}$, the timing of SF, ranging between $40\%$ and $60\%$ of the gait cycle; (2) $\theta_{SF}$, the targeted knee flexion angle ($40^\circ - 60^\circ$); (3) $k_{SF}$, the stiffness of SF ($0.4 - 1.8\,\mathrm{Nm/^\circ}$); (4) $k_{SE}$, the stiffness of SE ($0.35 - 0.6\,\mathrm{Nm/^\circ})$.

\subsection{Protocol}
For familiarization, after fitting and aligning the prosthesis to the user, they walk with the knee locked in an extended and stiff position to safely introduce the system. Once confident, the prosthesis is switched to a walking mode with a baseline configuration $x_{ref}$ with intermediate values in each dimension, that is, $t_{SF}=50\%$, $k_{SF}=1.3 Nm/^\circ$, $\theta_{SF}= 50^\circ$, and $k_{SE}=0.6Nm/^\circ$. Participants walk with the configuration $x_{ref}$ for multiple trials between 3-5 min. The whole session lasts 1 hour and breaks are provided
between trials.

Once the user is confident with the prosthesis, we construct the initial model based on their preferences over $N_{initial}$ randomly samples parameter pairs. Then, the user performs three optimization trials. Each trial aims to identify a user-preferred configuration using the pairwise comparison loop described in the previous section. In trial $i$, the algorithm runs until the convergence criteria is reached or the budget $N$ is depleted, and outputs a preferred configuration $\hat{x}_{N, i}$.

After the three optimization trials are completed, a single validation phase is conducted. During this phase, the three preferred configurations $\{\hat{x}_{N, i} \; | \; i \in \{1,2,3\}\}$, are compared each to three randomly selected candidates. The order of the configurations within each pair are randomized in order to prevent from any bias.

During validation, knee joint kinematics were recorded using the encoder integrated in the prosthetic knee for both the preferred configurations and randomly selected candidates. Mean knee angle trajectories were obtained by averaging joint angle signals over all recorded gait cycles. Swing and stance durations were derived from the gait phase variable, and corresponding asymmetry indices were computed to quantify temporal differences between the prosthetic and intact limbs. An asymmetry value of 0 indicates perfect symmetry, whereas negative values reflect longer loading on the prosthetic side. Such temporal asymmetry metrics are commonly used to characterize compensatory strategies and overall gait quality in transfemoral amputees~\cite{alili2023novel}. 

\begin{table}
\centering
\caption{Converged configurations obtained for each participant.
Each participant completed three trials, yielding preferred parameter vectors $\{\hat{x}_{N, i} \; | \; i \in \{1,2,3\}\}$ of the form
$\left[t_{\mathrm{SF}},\, k_{\mathrm{SF}},\, \theta_{\mathrm{SF}},\, k_{\mathrm{SE}}\right]$.
For participants AB1, AB2, and TF1, optimization was conducted over a discrete grid with five points per parameter, whereas TF2 was optimized in a continuous space.}
\label{tab:convergence_configs}
\renewcommand{\arraystretch}{1.15}
\begin{tabular}{c c c c c c}
\toprule
Subject & Config. & $t_{\mathrm{SF}}$ (\%) & $k_{\mathrm{SF}}$ (Nm/$^\circ$) & $\theta_{\mathrm{SF}}$ ($^\circ$) & $k_{\mathrm{SE}}$ (Nm/$^\circ$) \\
\midrule
\midrule
\multirow{3}{*}{AB1} & $\hat{x}_{N, 1}$ & 45 & 0.4 & 45 & 0.35 \\
 & $\hat{x}_{N, 2}$ & 50 & 0.4 & 40 & 0.35 \\
 & $\hat{x}_{N, 3}$ & 55 & 0.4 & 50 & 0.35 \\
\midrule
\multirow{3}{*}{AB2} & $\hat{x}_{N, 1}$ & 40 & 0.4 & 55 & 0.35 \\
 & $\hat{x}_{N, 2}$ & 40 & 0.4 & 45 & 0.35 \\
 & $\hat{x}_{N, 3}$ & 40 & 1.8 & 45 & 0.35 \\
\midrule
\multirow{3}{*}{TF1} & $\hat{x}_{N, 1}$ & 50 & 1.4 & 40 & 0.35 \\
 & $\hat{x}_{N, 2}$ & 40 & 0.4 & 40 & 0.35 \\
 & $\hat{x}_{N, 3}$ & 45 & 1.4 & 40 & 0.35 \\
\midrule
\midrule
\multirow{3}{*}{TF2} & $\hat{x}_{N, 1}$ & 53.2 & 1.0 & 40.0 & 0.6 \\
 & $\hat{x}_{N, 2}$ & 60.0 & 1.8 & 45.7 & 0.6 \\
 & $\hat{x}_{N, 3}$ & 43.4 & 0.6 & 44.8 & 0.5 \\
\bottomrule
\end{tabular}
\end{table}

\subsection{Human subject results}
\label{sec:human_results}

\textbf{\textit{EUBO-LineCoSpar}.} The algorithm is run with $N_{init} = 5$ pairs, $N=15$ iterations, and $N_{stop} =3$ iterations, ensuring a low number of user interactions in a simplified discrete setting to avoid user fatigue. Three participants participated in the study: one transfemoral amputee (TF1, age: 30 y.o., height: 180cm, weight: 65kg) and two able-bodied individuals wearing an adaptor (AB1 and AB2: age: 22.5$\pm$2y.o, height: 168$\pm$7cm, weight: 64.5$\pm$5kg). The experiments confirm users can consistently express reliable preferences across different prosthesis assistance configurations. As shown in Table~\ref{tab:convergence_configs}, each user converged to different set of parameters, $\{\hat{x}_{N, i} \; | \; i \in \{1,2,3\}\}$, specific to them. Moreover, they were able to recognize $\hat{x}_{N, i}$ during validation in $93\%$ of the cases. The algorithm converged in fewer than 15 iterations (10.3±2.5 iterations), with each trial lasting a total of 8.5±3.6 minutes, indicating the algorithm's high sample efficiency. 

For TF1, the optimized assistance of the CYBERLEG X-Leg prosthesis demonstrated improved gait symmetry compared to their personal prosthesis, particularly in swing and stance duration asymmetries (from -13.4\% to -7.6\% and from 6.2\% to 3.9\%, respectively).
\vspace{3pt}
\\
\textbf{BPE4Prost.} The algorithm is run with $N_{init} = 15$ pairs, $N=40$ iterations, and $N_{stop} =5$ iterations, to have maximum budget to explore the parameter space. One subject with transfemoral amputation participated in the study (TF2, age: $59$~years, $168$~cm, $68$~kg). The initial design lasted $6.8$ minutes, and each optimization trial $13.7 \pm 4.7$ minutes for $35.0 \pm 6.2$ iterations. 

\begin{figure}[h]
    \centering
    \includegraphics[width=0.49\textwidth]{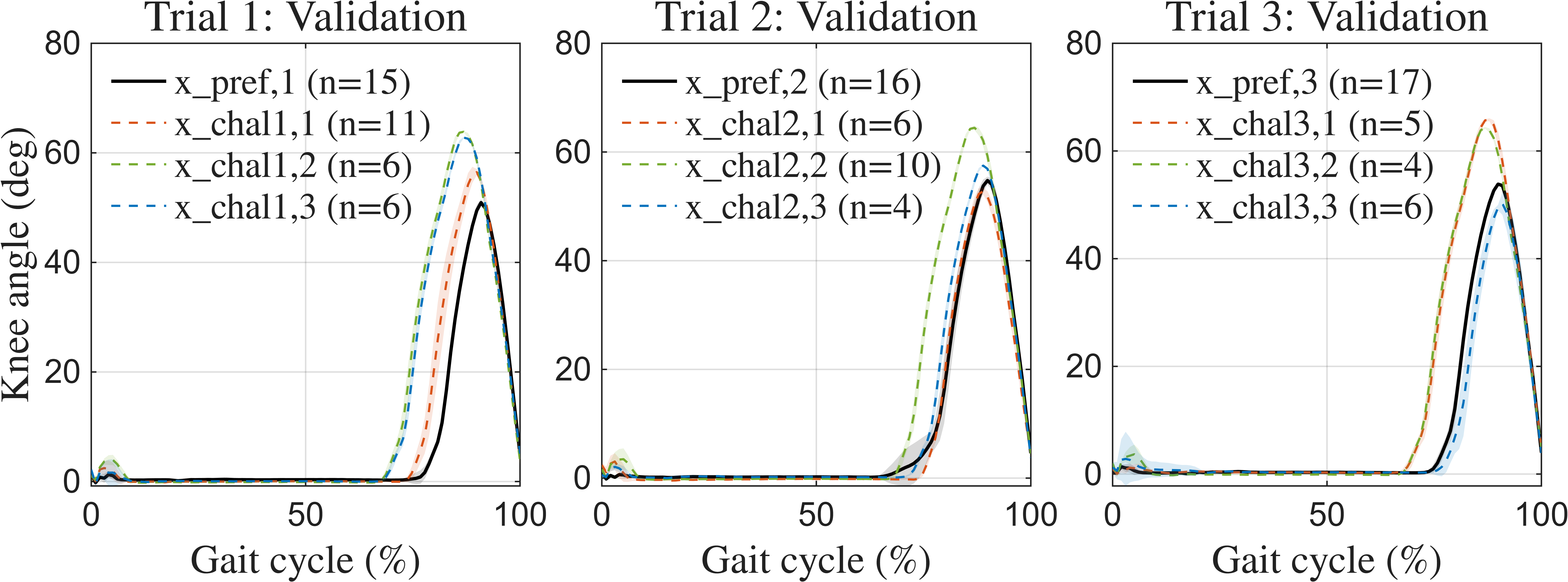}
    \vspace{-20pt}
    \caption{Comparison of knee angle trajectories for preferred solutions and their corresponding challengers. Each subplot displays the mean $\pm$ std of the knee angle over the normalized gait cycle (0–100\%) across steps. The number of gait cycles used to compute each mean trajectory is indicated.}
    \label{fig:knee_angle_comparison}
\end{figure}

As reported in Table~\ref{tab:convergence_configs}, in trials 1 and 2, TF2 converged to the same values for $k_{\mathrm{SE}}$, while $\theta_{\mathrm{SF}}$, $t_{\mathrm{SF}}$, and $k_{\mathrm{SF}}$ exhibited some variability (approximately $25\%$, $34\%$, and $57\%$ of their respective ranges). In trial 3, differences in  $t_{\mathrm{SF}}$ and $k_{\mathrm{SF}}$ were more pronounced. However, as shown in Figure~\ref{fig:knee_angle_comparison}, despite a $\sim 25\%$ variation in $\theta_{\mathrm{SF}}$ across trials, the maximum flexion angles reached for the preferred solutions remain similar (in average $51.0 \pm 0.5^\circ$, $55.0 \pm 0.6^\circ$, and $54.0 \pm 0.2^\circ$ for $\hat{x}_{N, 1}$, $\hat{x}_{N, 2}$, and $\hat{x}_{N, 3}$ respectively). 

During the validation process, TF2 correctly identified its preferred configuration in 67\% of trials against random challengers. As illustrated in Figure~\ref{fig:knee_angle_comparison}, for two of the three challengers where TF2 selected the challenger (i.e., $x_{\mathrm{chal2,1}}$ and $x_{\mathrm{chal3,3}}$) instead of $\hat{x}_N$, the mean maximum knee flexion angles were nearly identical to those of the corresponding preferred solutions, with values of $53.0 \pm 0.6^\circ$ and $51.0 \pm 0.3^\circ$, respectively. The third error in validation was for $x_{\mathrm{chal2,2}}$.

\begin{figure}[h]
    \centering
    \includegraphics[width=0.45\textwidth]{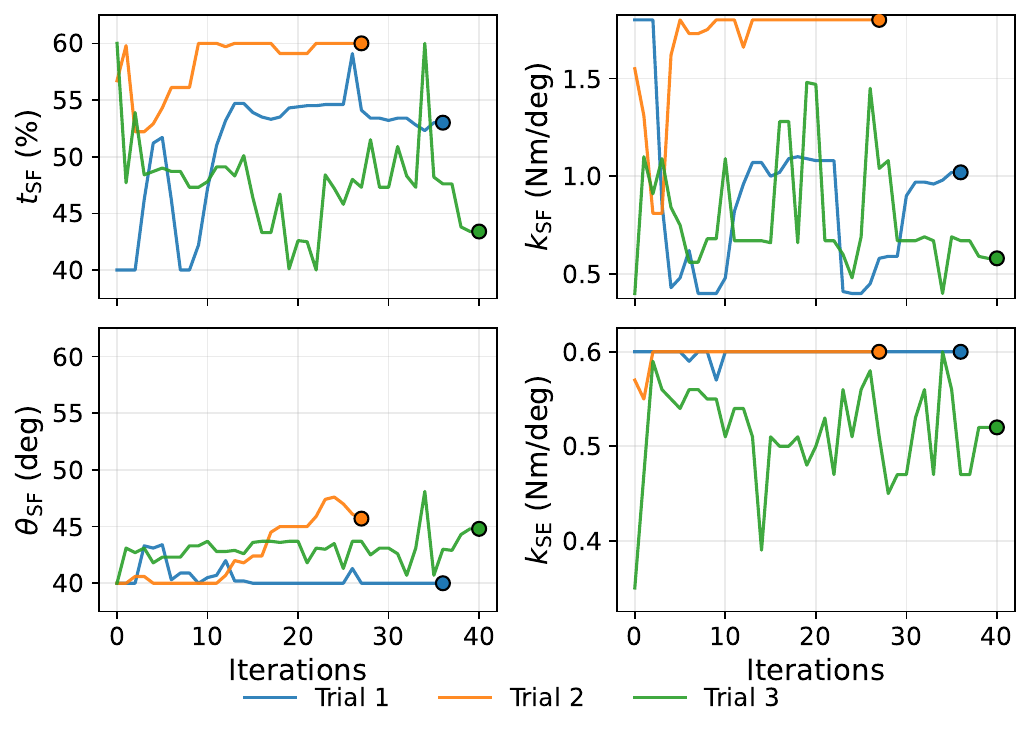}
    \vspace{-20pt}
    \caption{Evolution of the estimated optimal configuration for TF2, $\hat{x} = \arg\max_x \mu_{N}(x)$, for the three trials, shown for each dimensions. The final estimate for each trial is highlighted. Trials 1 and 2 converge rapidly in most of the dimension, whereas Trial 3 exhibits increased variability, likely due to noisy user feedback.}
    \label{fig:TF2_estimate_cv}
\end{figure}

Figure~\ref{fig:TF2_estimate_cv} illustrates the evolution of the model’s estimate of the user-preferred configuration, $\hat{x}_N$, across iterations for each trial. In Trial~1 (blue curve), the estimated preference remains largely stable from iteration~15 onward, with variations below 15\% in all dimensions. A temporary deviation is observed between iterations~24 and~30, primarily affecting $k_{\mathrm{SF}}$, which is likely attributable to a noisy user response. In Trial~2 (orange curve), the estimate stabilizes from iteration 15 onward, with rapid convergence achieved by iteration~27. In contrast, Trial~3 exhibits sustained variability in the estimated preference beyond iteration~15, suggesting increased noise in the user feedback, potentially caused by user fatigue.
\section{Discussion}
\label{sec:discussion}
Together, the simulation and human subject results provide a detailed picture of how preference-based optimization behaves in this setting. 

\subsection{User preference in prosthesis control tuning}
\textbf{Tuning Time Reduction.}
FSM impedance controllers are widely used for transfemoral prostheses, but tuning their parameters across ambulation modes remains time-consuming. Simon \textit{et al.}~\cite{simon2014configuring} showed that even after parameter reduction, novice users required 2.5–5 hours for personalization, corresponding to 46–92 minutes to tune four parameters. In contrast, our \textit{BPE4Prost} framework demonstrates the potential to substantially reduce tuning time. In our experiments, optimization was conducted over a 40-iteration budget, with an initial design phase of approximately 7 minutes followed by around 15 minutes of optimization. However, convergence toward satisfactory configurations was often observed earlier, suggesting that the process could be further shortened in practice. Moreover, \textit{EUBO-LineCoSpar} exhibited rapid convergence in the discrete setting, requiring less than 10 minutes for initial design and optimization. Overall, such a framework could enable fast identification of an acceptable configuration, which could then be refined by a prosthetist if necessary. This potential is further supported by the observed improvements in swing and stance duration asymmetries, indicating that preference-driven tuning can enhance prosthetic performance and gait quality in clinical applications.
\vspace{3pt}
\\
\textbf{User preference consistency.} Previous studies have applied BPE to prosthetic and exoskeleton control, tuning 12~\cite{alili2023novel}, 4~\cite{lee2023user} and 2~\cite{arens2025preference} parameters. Consistent with our findings, all reported substantial inter-subject variability, highlighting the individualized nature of preference, with users correctly identifying their preferred configurations in approximately $90\%$ of pairwise comparisons. Alili \textit{et al.}~\cite{alili2023novel} further investigated whether user preference could be explained by biomechanical variables, assessing correlations with margin of stability and gait phase symmetry, but found no significant association with those gait metrics. 

For HILO, incorporating user preference into the cost function can enhance personalization by ensuring that the selected assistance reflects both physiological and perceptual factors, such as comfort and perceived effort. However, it remains unclear which biomechanical variables drive these preferences. Ongoing work therefore seeks to systematically integrate additional gait parameters (e.g., gait smoothness, stance duration) to better capture user-driven performance and refine prosthesis optimization. Identifying the most relevant variables remains a critical step toward more effective, preference-informed control.
\vspace{3pt}
\\
\textbf{User perceptual limits.} During the optimization sessions, users occasionally struggled to discriminate between closely spaced configurations, particularly toward the end of a trial when the algorithm focused on exploiting high-utility regions. This underscores the importance of designing queries that respect human perceptual limits. One effective strategy is to enforce minimum-distance constraints in the acquisition function, ensuring that suggested configurations differ by at least the Just-Noticeable Difference (JND) for each parameter. The concept of JND has been explored in studies by Medrano \textit{et al.}~\cite{medrano2020methods}, providing guidance on perceptible parameter variations. Similarly, Arens \textit{et al.}~\cite{arens2025preference} incorporated a JND constraint into their Upper-Confidence Bound acquisition function to improve preference elicitation. Extending such approaches in future work could enhance both user comfort and the reliability of preference models.

\subsection{Model design}
\textbf{Acquisition function.} Sun \textit{et al.}~\cite{sun2024individual} proposed a BPE relying on random pairwise sampling and subjective preference feedback to infer personalized prosthesis control strategies. However, their method selects candidates randomly and adjusts only one parameter per comparison, which can result in inefficient queries and slow convergence. In contrast, our framework leverages the EUBO acquisition function to guide candidate selection efficiently across the multidimensional parameter space (Figure~\ref{fig:simulations}). By explicitly targeting high-utility regions, EUBO reduces uninformative comparisons and accelerates convergence.
\vspace{3pt}
\\
\textbf{Optimization budget.} Choosing an appropriate number of iterations is critical to balance exploration and exploitation. With \textit{EUBO-LineCoSpar}, the limited evaluation budget made it difficult to determine whether the optimization converged to a local or global optimum. By increasing the budget in \textit{BPE4Prost}, we were able to observe the model behavior more clearly over a larger set of pairwise comparisons. This allowed for an initial exploration of the parameter space, followed by a more localized exploitation phase, providing insights into the convergence dynamics of the preference model.
\vspace{3pt}
\\
\textbf{Kernel hyperparameter priors.} 
Simulations highlighted the importance of placing informative priors on kernel hyperparameters. Because the latent utility is never directly observed, the GP's hyperparameters are often weakly identified, which can lead to unrealistically large lengthscales or output scales. Such large values results in a near constant posterior and flatten the acquisition landscape, impairing effective exploration. Incorporating reasonable priors on the GP's hyperparameters stabilizes the surrogate model and improves acquisition reliability.
\vspace{3pt}
\\
\textbf{Heteroscedastic and time-aware noise.} 
Both experiments suggest that the noise in user responses is not constant across sessions. In particular, variations in the consistency of preferences were observed between trials (see Figure~\ref{fig:TF2_estimate_cv}), indicating that the effective noise level may evolve over time. This highlights the importance of explicitly tuning or modeling the noise parameter. Allowing the noise scale to vary with iteration or context (e.g., session time or recent effort) can capture fatigue-driven variability. A simple approach is to put a prior on $\lambda$ and re-estimate it online; more structured variants include input-dependent noise models or kernels that include a time coordinate to account for non-stationary preferences. These extensions can reduce abrupt belief shifts and stabilize acquisition decisions late in a session.

\section{Conclusion}
\label{sec:conc}


This paper investigates the use of Bayesian Preference Elicitation for tuning an active transfemoral prosthesis. We introduce a Human-In-the-Loop Optimization framework that leverages pairwise user preferences and the EUBO acquisition function to enable fast and personalized adjustment of the control parameters that define knee assistance. By reducing the time and effort required for manual tuning, the proposed approach addresses key limitations of current prosthesis control procedures.

We presented preliminary experimental results obtained in real-life conditions, considering both a discretized and a continuous parameter space. In the discrete setting, results showed that users were able to express consistent preferences, allowing the algorithm to converge toward configurations that provided improved perceived assistance. In the continuous setting, similar trends were observed: the algorithm effectively balanced exploration and exploitation, leading to an estimated utility landscape that users could distinguish from randomly selected configurations. After 15 to 20 iterations, it already seems that the model can provide a good estimation of user preference. 

Ongoing work aims to extend these results through experiments with a larger number of participants, in order to better characterize inter-individual variability. In parallel, additional biomechanical measurements are being collected to gain deeper insight into the underlying factors that define user-specific preferences and to further inform the personalization of prosthesis control.

\section{Acknowledgements}
The authors thank the Amptraide organization for helping us to find participants and their interest in developing this work. A particular thanks to D.J. and M.J. for their participation in the experiments.

\section{Funding}
This work was supported by BOSA FOD AI-Driven Wearable Robotics for Healthcare (Aidwear) and the Strategic Research Program Exercise and the Brain in Health and Disease: The Added Value of Human-Centered Robotics, Vrije Universiteit Brussel, Belgium. 

Wouter Koppen is supported by the Flemish Government under the “Onderzoeksprogramma Artificiële Intelligentie (AI) Vlaanderen. His research is financially supported by the Research Foundation-Flanders (FWO) (G092923N). 

Sebastian Rojas Gonzalez acknowledges support by FWO (Grant number 1216021N) and the Belgian Flanders AI Research Program. 

Eligia Alfio is a doctoral fellow of the FWO (Grant number 1SA1W26N).

\bibliography{bibliography}
 
\end{document}